# Optimization of an Augmented R-CUBE mechanism for Cervical Surgery

Terence Essomba[1][0000-0002-8828-6802], Yu-Wen Wu[1], Abdelbadia Chaker[2] and Med Amine Laribi[2][0000-0003-0797-7669]

[1] National Central University, Taoyuan City 32001, Taiwan
[2] Department of GMSC, Pprime Institute CNRS, ENSMA, UPR 3346, University of Poitiers, France
tessomba@cc.ncu.edu.tw

**Abstract.** In some surgical operations targeting the spine, it is required to drill cavities in the vertebrae for the insertion of pedicle screws. A new mechanical architecture is proposed for this application. It is based on an augmented version of the full translational R-CUBE mechanism, with improved linkages to implement additional rotational motion. Using this concept, a mechanism presented with a 3T2R motion that is required for the manipulation of the surgical drill. It is mainly composed three stages: one translational, one transmitting and one rotational. Their respective kinematic and velocity models are separately derived, then combined. Based on the drilling trajectories obtained from a real patient case, the mechanism is optimized for generating the highest kinematic performances.

## 1 Introduction

Cervical arthrodesis is a surgical procedure that fuses or consolidates at least two vertebrae technique using a spinal implant. It relieves the nerves or spinal cord compression while stabilizing the spine by restoring proper vertebral spacing.

Accessing the vertebrae for implant fixation can be done through anterior (front) or posterior (back) approaches, chosen based on pathology, lesion extent, and surgical team preferences. [1]. Various fixation techniques exist, differing in the anatomical anchorage for the spinal implant screw(s) required to stabilize the spine [2]. Precise screw insertion requires detailed anatomical knowledge of each vertebra to be drilled, including entry points, direction, diameter, and depth. During surgery, accurate positioning of the surgical motor and drilling tool along the optimal trajectory is critical.

The pedicle sight procedure can be divided into three stages, each with distinct tool mobility requirements. The positioning of the surgical tool, the adjustment of its orientation to align with the optimal predefined drilling trajectory, and the drilling act along the optimal trajectory. The procedure is summarized in **Fig. 1**.

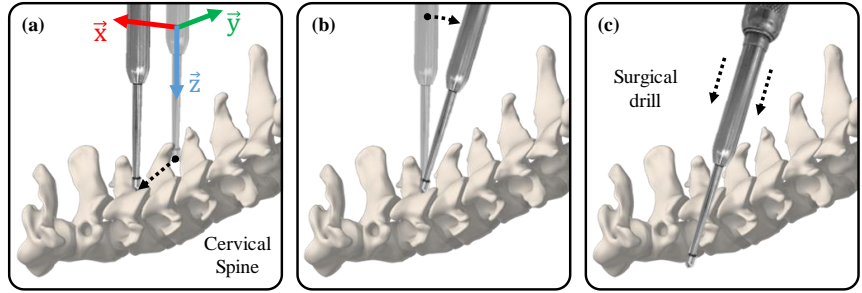

**Fig. 1.** Cervical drilling procedure: (a) Linear positioning of the drill tip, (b) orientation of the drilling axis and (c) longitudinal motion for cavity drilling.

### 1.1   Robotic Spinal Surgery

In Cervical arthrodesis context, integrating a robotic system in the operating room enhances precision and repeatability of the surgical procedure, enabling finer drillings and increased screw stability. Most of robot-assisted orthopedic procedures [3], listed in the literature, focus on pedicle screws implantation, at 56%, total hip arthroplasty, i.e., replacement of the hip joint with a prosthesis, followed by total or partial knee arthroplasty.

Existing robotic devices can be classified into three categories: First, autonomous robots, i.e., THINK Surgical (USA), [4], teleoperated robots, i.e., Da Vinci robot from Intuitive Surgical (USA) [5], and comanipulated robots [6], i.e., ROSA Spine (Medtech, France) based in Montpellier [7], Mazor X Stealth (Medtronic,United States) [8] and and ExcelsiusGPS (Globus Medical, United States) [9].

### 1.2   Augmented translational mechanism

Among parallel mechanical architectures, some tripod mechanisms with purely translational motion can be very useful in some applications. The world-known Delta robot is the was invented and patented in 1990 [10] and was subject to countless studies. Other specifically named mechanisms can be mentioned like the Orthoglide [11], the Tripteron [12], the R-CUBE [13] or the Star-Link mechanism [14]. And others can be quoted as well: the 3-UPU [15], the 3-PRC [16], the 3-PUU [17], the 3-CPU [18] or the 3-CUR [19]. Augmented translational mechanisms have modified leg architectures in order to generate additional DoF. As a result, higher mobility can be achieved without adding more legs causing the associated issues mentioned above. Very few instances can be found. An augmented version of the Tripteron was reported with six DoF but only three legs [20]. The Orthoglide was also augmented to the same type of improvement [21].

The R-CUBE has been intensively studied due to its potential as a simple architecture and its augmentation into a 3T3R motion mechanism was proposed [22]. The contribution of this study is to adept this augmented translational mechanism for cervical surgery. A modified version with 3T2R motion is specifically proposed, analyzed and optimized for this application. The next section will present the adapted architecture and its kinematic analysis. The dimensional optimization of the mechanism based on a real case drill trajectories is conducted in Section 3. The conclusion is provided in the last section.

## 2    Analysis of the AR-CUBE mechanism

The AR-CUBE mechanism is an augmented version of the R-CUBE that is sought to be adapted and optimized for cervical surgery. Angular DoF are added to the mechanism by modifying the planar distal linkages of the classical R-CUBE. These linkages transmit an angular motion to an added spherical mechanism.

### 2.1    Kinematic model

As illustrated on **Fig. 2**, the AR-CUBE is composed of three stages: a translational stage (TLS), a transmitting stage (TMS) and a rotational stage (RTS). The two first stages are shown on **Fig. 2**-(a) and the RTS is detailed on **Fig. 2**-(b). The TLS is directly coming from the classical R-CUBE and directly controls the three-DoF linear motion. The TMS comes from the passive distal linkages that are modified to now transmit a two-DoF angular motion to the RTS. In the present application, only two angular DoF are required, so only two distal linkages are modified, the third remains as original in the R-CUBE.

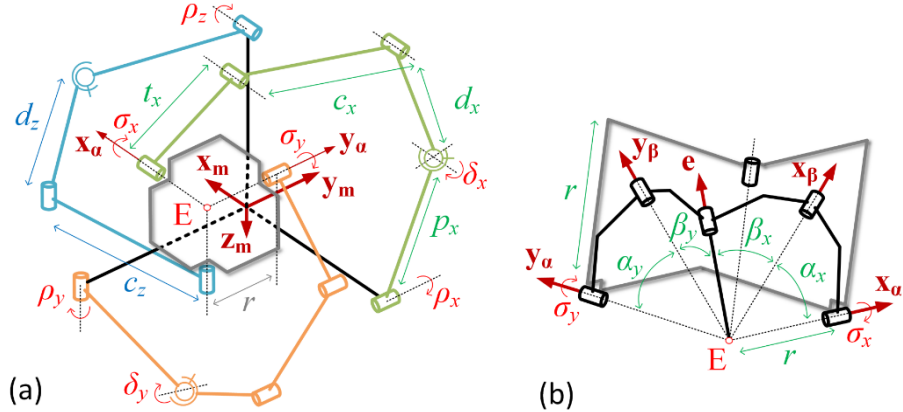

**Fig. 2.** Kinematic drawing of the AR-CUBE mechanism adapted for cervical surgery. (a) TLS and TMS, (b) RTS.

The origin of the mechanism is represented by the point $O_m$ and its end effector by the point E, which is the tip of the surgical drill of coordinates $x$, $y$ and $z$. The reference frame associated with the mechanism is noted $R_m = \{O_m, \mathbf{x_m}, \mathbf{y_m}, \mathbf{z_m}\}$. To conduct the mechanism analysis in a generalized form for each leg, a coordinates system $\{\mathbf{u}, \mathbf{v}, \mathbf{w}\}$ is built where $\mathbf{u} = \mathbf{x}$, $\mathbf{y}$ or $\mathbf{z}$, and $\mathbf{v}$ and $\mathbf{w}$ are the following unit vectors. There a total of five active joints: three proximal and two distal labelled as points $P_u$ and $D_u$. Active joints $P_u$ rotate around axis **v** with angles $\rho_u$ and $D_u$ around axis **u** with angles $\delta_u$. The inverse linear kinematics of the mechanism is exclusively driven by the TLS and directly calculated for each leg as follows:

$$\rho_u = -\sin^{-1}\left(\frac{u+f_u+r}{p_u}\right), \tag{1}$$

Where $u$ is one linear coordinate $x$, $y$ or $z$.

The TMS ensures the transmission of the angular motion from the distal joint to the TRS. It is made of a classical four-bar linkage with an input link length $d_u$, a coupling link length $c_u$ and a transmitting link length $t_u$. The joint variable $\delta_u$ is measured between the link $d_u$ and axis **w**. It allows controlling the output angle $\sigma_u$ between link $t_u$ and axis **w**. The inverse kinematics is found by writing the following constraint equation:

$$c_u^2 = (v - d_u \sin\delta_u + t_u \sin\sigma_u)^2 + (w + d_u \cos\delta_u - t_u \cos\sigma_u - p_u \cos\rho_u)^2. \tag{2}$$

The equation is solved by isolating the joint distal joint variable as follows:

$$\delta_u = \text{ATAN2}\left(\frac{B_u}{\sqrt{A_u^2+B_u^2}}, \frac{A_u}{\sqrt{A_u^2+B_u^2}}\right) \pm \cos^{-1}\left(\frac{C_u}{\sqrt{A_u^2+B_u^2}}\right), \tag{3}$$

Where $A_u$, $B_u$ and $C_u$ are functions of the mechanism variables and parameters.

A five-bar spherical linkage is selected for the TRS [23]. Its function is to control the orientation of the surgical drill axis with two DoF. The output variables of the TMS directly transmit the distal active joint motions to the input variables of the TR, so the latter is equal to $\sigma_u$. The two legs of the TRS connect with the end effector of the AR-CUBE. Each leg is made of one proximal and one distal spherical link of dimension $\alpha_u$ and $\beta_u$ respectively and their proximal and distal joint axes are noted $\mathbf{u}_\alpha$ and $\mathbf{u}_\beta$, with u = x or y only as there are only two legs. The end effector is given by the joint axis **e** and its orientation is measured by the angles $\psi$ and $\theta$. The constraint equation of the RTS is written as follows:

$$\mathbf{u}_\beta \cdot \mathbf{e} = \cos\beta_u \tag{4}$$

Expanding Eq. (4) leads to an equation with the same expression as Eq. (3) to calculate the RTS input variables $\sigma_u$. However, functions $A_u$, $B_u$ and $C_u$ will have different expressions.

### 2.2 Velocity model

The velocity model of the AR-CUBE is calculated using differential kinematics. The kinematic model of the different stages of the mechanism are differentiated with the time. It is sought to obtain the relationship between the proximal and distal input joint velocities $\dot{\boldsymbol{\rho}}$ and $\dot{\boldsymbol{\delta}}$ and the linear and angular end effector velocities **v** and **ω**.

The linear velocity model of the mechanism is directly obtained by differentiating the forward linear kinematics of the TLS, which leads to the following matrix form:

$$\mathbf{v} = \text{diag}(p_u \cos \rho_u)\, \dot{\boldsymbol{\rho}} = \mathbf{J}_{LV} \cdot \dot{\boldsymbol{\rho}}, \tag{5}$$

Where **v** is the end effector linear velocity and $\dot{\boldsymbol{\rho}}$ is the proximal joint velocities.

For the angular velocity model, the TMS and RTS are considered. First, the analysis of the TMS give the relationship between the distal joint velocities $\dot{\boldsymbol{\delta}}$ and its output velocities $\dot{\boldsymbol{\sigma}}$, which corresponds to the input velocities of the RTS. The TMS constraint Eq. (2) is differentiated with the time and arranged in a matrix form considering both legs as follows:

$$\begin{bmatrix} F_x & 0 \\ 0 & F_y \end{bmatrix} \dot{\boldsymbol{\delta}} + \begin{bmatrix} G_x & 0 \\ 0 & G_y \end{bmatrix} \dot{\boldsymbol{\rho}} = \begin{bmatrix} 0 & D_x & E_y \\ E_y & 0 & D_y \end{bmatrix} \mathbf{v} + \begin{bmatrix} H_x & 0 \\ 0 & H_y \end{bmatrix} \dot{\boldsymbol{\sigma}}, \tag{6}$$

Where functions $D_u$ to $H_u$ are expressions of the mechanism variables and parameters. It is then written in a simplified form as:

$$\mathbf{J}_\rho \dot{\boldsymbol{\rho}} + \mathbf{J}_\delta \dot{\boldsymbol{\delta}} = \mathbf{J}_{AV}\mathbf{v} + \mathbf{J}_\sigma \dot{\boldsymbol{\sigma}}. \tag{7}$$

For the RTS, the constraint Eq. (4) is differentiated with the time and written in a matrix as below:

$$\begin{bmatrix} x_\alpha \times x_\beta \cdot e & 0 \\ 0 & y_\alpha \times y_\beta \cdot e \end{bmatrix} \dot{\boldsymbol{\sigma}} = \begin{bmatrix} (x_\beta \times e)^T \\ (y_\beta \times e)^T \end{bmatrix} \boldsymbol{\omega} \tag{8}$$

And simplified as:

$$\mathbf{J}_{AS}\dot{\boldsymbol{\sigma}} = \mathbf{J}_{AP}\boldsymbol{\omega}. \tag{9}$$

To complete the angular velocity model, Eq. (7) is substituted with Eq. (9) as follows:

$$\mathbf{J}_\rho \dot{\boldsymbol{\rho}} + \mathbf{J}_\delta \dot{\boldsymbol{\delta}} = \mathbf{J}_{AV}\mathbf{v} + \mathbf{J}_\sigma (\mathbf{J}_{AS}^{-1}\mathbf{J}_{AP})\boldsymbol{\omega}. \tag{10}$$

Using Eq. (5) and (7), the entire velocity model of the 3T2R AR-CUBE can be written as:

$$\begin{bmatrix} \mathbf{J}_{LV} & \mathbf{0}_{3\times 3} \\ \mathbf{J}_{AV} & \mathbf{J}_\sigma(\mathbf{J}_{AS}^{-1}\mathbf{J}_{AP}) \end{bmatrix} \begin{bmatrix} \mathbf{v} \\ \boldsymbol{\omega} \end{bmatrix} = \begin{bmatrix} \mathbf{I}_{3\times 3} & 0 \\ \mathbf{J}_\rho & \mathbf{J}_\delta \end{bmatrix} \begin{bmatrix} \dot{\boldsymbol{\rho}} \\ \dot{\boldsymbol{\delta}} \end{bmatrix}. \tag{11}$$

## 3    Surgical-Oriented Optimization

The AR-CUBE mechanism is optimized for cervical surgery based on trajectories collected from a real patient case. The objective is to identify the mechanism dimensions that will generate the highest possible kinematic performances.

### 3.1 Index formulations

It is aimed for the AR-CUBE to have the highest possible kinematic performance along a series of drilling trajectories. Therefore, one constraint is the reachability of the trajectories within the mechanism workspace and one criterion is the kinematic performance. The workspace is defined based on a set of directions for the drilling trajectories as depicted in **Fig. 3**. The digital model of the illustrated cervical spine is obtained from the CT-scan of a patient. The trajectories are equally discretized in a certain number of points with linear coordinates for the position of the drill tip and angular coordinates for the orientation of the drill axis. It is noted that the orientation remains constant on the same trajectory. The coordinates of the reference frame $R_m$ representing the mechanism position and the coordinates of the trajectory points are all given the reference frame $R_w$.

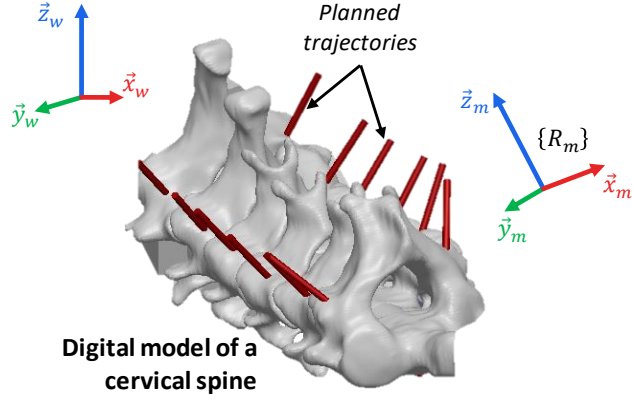

**Fig. 3.** Drilling trajectories representation on the digital model of the cervical spine.

The reachability of each point must be ensured to validate the mechanism workspace constraint. The associated conditions are determined by existence of a solution to Eq. (1) for the TLS and Eq. (3) for the TMS and RTS for all active joint variables. For each point, it is checked if the TLS can reach its linear coordinates $x$, $y$ and $z$. Then, it is verified that the RTS can reach its angular coordinates $\psi$ and $\theta$. And finally, it is tested if the TMS can establish a feasible connection between the TLS and the RTS. A function $w_L$ is defined as a local index for the reachability of a given point of coordinates $(x, y, z, \psi, \theta)$. The function value is 1 is the point is reachable and 0 if not. The reachability of all points on all trajectories is then defined as follows:

$$w_G = \prod_{i=1}^{n} w_L(x, y, z, \psi, \theta), \tag{12}$$

Where $n$ is the number of points on all the trajectories. The global reachability $w_G$ will return 1 if all points are reachable and 0 if at least one point is not. If this condition is validated for all points, the mechanism candidate is evaluated using the kinematic performance as a criterion.

The kinematic performance is often referred to as dexterity when it is estimated over a mechanism workspace. In the present case, it is defined as the average of the local dexterity calculated at all points of the trajectories. For each point, the condition number of all Jacobian matrices (a total of seven) is calculated and the worst case (which corresponds to the highest value) is retained as the local index as follows:

$$\eta_L = \max(\kappa(\mathbf{J_{LV}}) \quad \kappa(\mathbf{J_{AV}}) \quad \kappa(\mathbf{J_\sigma}) \quad \kappa(\mathbf{J_{AS}}) \quad \kappa(\mathbf{J_{AP}}) \quad \kappa(\mathbf{J_\rho}) \quad \kappa(\mathbf{J_\delta})). \tag{13}$$

The global dexterity is then calculated as:

$$\eta_G = \frac{1}{n}\sum_{i=1}^{n} \eta_L(x, y, z, \psi, \theta). \tag{14}$$

Considering the mechanism workspace and dexterity, the present optimization problem can be formulated as follows:

$$\begin{aligned}&\text{Minimize } \eta_G \\ &\text{Subject to } \eta_G = 0\end{aligned} \tag{15}$$

Each mechanism candidate is represented by a vector which terms are the mechanism dimensions. They are all listed in **Table 1**.

Table 1. List of mechanism link and parameter dimensions to optimize.

| Mechanism stages | Link and parameters | Dimensions | Number |
|---|---|---|---|
| TLS | Base position | $m_u$ | $x,y$ and $z$ |
| | Frame link | $f_u$ | $x,y$ and $z$ |
| | Proximal link | $p_u$ | $x,y$ and $z$ |
| TMS | Distal link | $d_u$ | $x,y$ and $z$ |
| | Coupling link | $c_u$ | $x,y$ and $z$ |
| | Transmitting link | $t_u$ | $x$ and $y$ only |
| RTS | Radius | $r$ | Unique |
| | Transmitting offset | $o_u$ | $x$ and $y$ only |
| | Proximal link | $\alpha_u$ | $x$ and $y$ only |
| | Distal link | $\beta_u$ | $x$ and $y$ only |

### 3.2  Optimization results

The optimization is carried out using a Genetic Algorithm (GA). It is program on Matlab software to manage and manipulate a population of mechanism candidates randomly defined by a set dimension. Each is evaluated by a fitness function based on the formulation of Eq. (15). The GA is executed for a population of 400 candidates over 300 generations. The convergence condition is set to $10^{-5}$. Each drilling trajectory is equally discretized into 30 points, which represents around 1mm of distance between two successive points. The optimum identified mechanism has a set of dimensions listed in **Table 2**.

**Table 2.** Link and parameter dimensions of the optimum mechanism.

| Parameters | x | y | z |
|:---:|:---:|:---:|:---:|
| $m_u$ | 31.8694 | 11.9589 | 192.0769 |
| $f_u$ | 123.1147 | 116.9719 | 56.7787 |
| $p_u$ | 134.099 | 149.9886 | 99.8766 |
| $d_u$ | 79.2596 | 99.9402 | 53.5671 |
| $c_u$ | 139.8911 | 106.8319 | NA |
| $t_u$ | 70.9301 | 65.3946 | NA |
| $r$ | 70.4466 (unique) | | |
| $o_u$ | -19.2229 | 43.2917 | NA |
| $\alpha_u$ | 52.6402 | 66.446 | NA |
| $\beta_u$ | 50.0038 | 62.1208 | NA |

To facilitate the evaluation, the dexterity (after optimization) is inverted to obtain a ration from 0 to 100% (the higher, the better). Over the while moving along the cervical drilling trajectories, it has average of 40.25%, with local values ranging from 20.26% to 66.48% as displayed on the graphic of **Fig. 4**. The discontinuities observed are caused by the change from one drilling trajectory to another, which is simulated instantaneously in the present study.

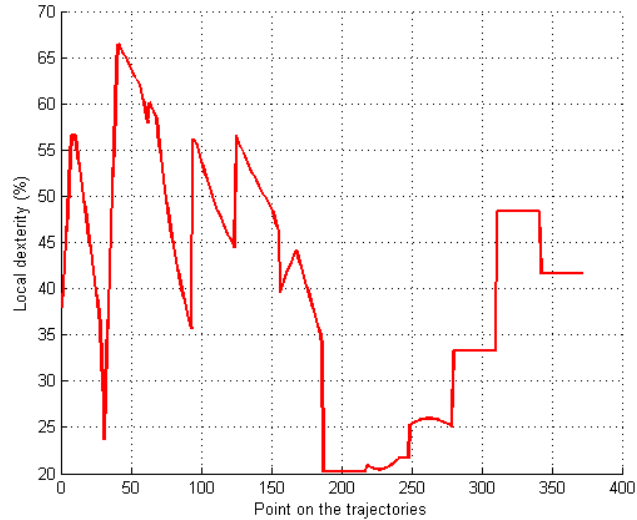

**Fig. 4.** Evolution of the local dexterity along the cervical drilling trajectories.

## 4     Conclusion

In the present student, an augmented mechanism is optimized for cervical surgery. The mechanism is based on a translational mechanism, namely, the R-CUBE. Architectural improvements have been made to increase its number of DoF without adding more legs in parallel. The new mechanism obtained generates a 3T2R motion and is called AR-CUBE. Two legs made of four-bar linkages are used to transmit the two additional DoF to a five-bar spherical linkage to achieve two angular DoF. The analysis of the TLS allows finding the linear kinematics of the AR-CUBE. The angular kinematics relies on the analysis of the TRS in series with the RTS. The velocity models of these three mechanism stages are combined using a total of seven Jacobian matrices, which are used for the definition of the AR-CUBE dexterity. The mechanism is then optimized in terms of dexterity along a series of cervical drilling trajectory issues from a real patient case. A GA is programed and executed to identify the 23 dimensions of the AR-CUBE that generates an average dexterity of around 40% and maximum 60%.

10      T. Essomba, Y.-W. Wu and M.A. Laribi